\documentclass[nopreprintline]{elsarticle}

\usepackage[margin=1in]{geometry}
\usepackage[modulo]{lineno}
\usepackage{comment}
\usepackage[hidelinks]{hyperref}

\usepackage{amsmath}
\usepackage{amsmath}
\usepackage{amssymb}
\usepackage{gensymb}
\usepackage{amsthm}
\usepackage{array,xcolor,colortbl}
\usepackage{multirow}
\usepackage{caption}
\usepackage{subcaption}
\usepackage{graphicx}
\usepackage{makecell}
\usepackage{float}
\usepackage{setspace}
\usepackage[figuresright]{rotating}
\usepackage{adjustbox}
\usepackage{lineno}

\definecolor{mygray}{gray}{.9}
\linenumbers

\begin{document}
\nolinenumbers

\begin{frontmatter}


\title{Multi-Modal Fusion of In-Situ Video Data and Process Parameters for Online Forecasting of Cookie Drying Readiness}

%

\author[a]{Shichen Li}
\ead{shichen8@illinois.edu}


\author[b,a]{Chenhui Shao\corref{corrauth}}
\ead{chshao@umich.edu}
\cortext[corrauth]{Corresponding author}

\address[a]{Department of Mechanical Science and Engineering, University of Illinois at Urbana-Champaign, Urbana, IL 61801, United States}


\address[b]{Department of Mechanical Engineering, University of Michigan, Ann Arbor, MI 48109, United States}

\begin{abstract}

Food drying is essential for food production, extending shelf life, and reducing transportation costs. Accurate real-time forecasting of drying readiness is crucial for minimizing energy consumption, improving productivity, and ensuring product quality. However, this remains challenging due to the dynamic nature of drying, limited data availability, and the lack of effective predictive analytical methods. To address this gap, we propose an end-to-end multi-modal data fusion framework that integrates in-situ video data with process parameters for real-time food drying readiness forecasting. Our approach leverages a new encoder-decoder architecture with modality-specific encoders and a transformer-based decoder to effectively extract features while preserving the unique structure of each modality. We apply our approach to sugar cookie drying, where time-to-ready is predicted at each timestamp. Experimental results demonstrate that our model achieves an average prediction error of only 15 seconds, outperforming state-of-the-art data fusion methods by 65.69\% and a video-only model by 11.30\%. Additionally, our model balances prediction accuracy, model size, and computational efficiency, making it well-suited for heterogenous industrial datasets. The proposed model is extensible to various other industrial modality fusion tasks for online decision-making.

\end{abstract}


\end{frontmatter}

\section{Introduction}

Drying is a fundamental process in the food industry that plays a critical role in both food production and preservation. By removing moisture, it transforms raw ingredients into their final, consumable forms while enhancing texture, flavor, and structural integrity~\cite{floros2010feeding}. However, food drying is a highly time- and energy-intensive process which accounts for 15\% of energy consumption in U.S. industrial processes~\cite{adnouni2023computational}. As a result, advancing drying technologies and improving product quality are key strategies for minimizing waste and enhancing energy efficiency~\cite{farzad2021drying}.

The quality of food drying is influenced by multiple factors, including environmental process parameters and the intrinsic properties of food samples~\cite{joardder2015porosity, arslan2024assessing, omolola2017quality, defraeye2017impact}. Prior research has shown that the inherent variability of food samples leads to inevitable process variances, making it difficult to achieve consistent drying outcomes ~\cite{li2025uncertainty}. To address such issues, process monitoring is essential for ensuring product quality and consistency~\cite{tian2023weldmon}. Due to the diverse factors influencing drying and their complicated interactions, effective monitoring must rely on accurate measurements and predictive modeling to characterize, understand, and control drying processes~\cite{schmitt2020predictive}. These requirements have led to two primary strategies: hardware-based sensing and data-driven modeling.

Hardware-based improvements use specialized sensors (e.g., near-infrared (NIR) spectroscopy) to directly measure product quality attributes~\cite{li2021human}. However, high device costs, complex setups, and parameter-specific designs limit their industrial adoption~\cite{qu2024development}. As such, data-driven modeling, which leverages process data to extract quality indicators without requiring expensive sensors or device-specific calibrations, emerges as a more attractive solution~\cite{zhao2023ai, jia2024hybrid}. Moving beyond traditional linear response surface methodologies, machine learning (ML) models excel at capturing the nonlinear and complex relationships inherent in food drying processes~\cite{meng2024meta, meng2023explainable}.

Various studies have explored data-driven monitoring in food drying~\cite{zhao2021effects, el2023novel}.  Many existing methods rely on offline evaluations or pre-calibrated models, limiting their ability to capture real-time interactions between sample variability and quality attributes~\cite{mishra2023development}. On the other hand, online monitoring aims to provide real-time assessments and timely interventions based on in-situ measurements~\cite{aghbashlo2014measurement}. One critical yet underexplored issue is predicting “drying readiness”—the remaining time to reach the optimal endpoint, which is vital for preventing over-drying, under-drying, and curbing energy consumption~\cite{chen2025reinforcement}.

Current in-situ monitoring methods hold strong potential for improving process visibility, particularly through computer vision (CV) techniques that can capture rich visual information without interrupting the drying process~\cite{shang2023defect}. Unlike manual measurements, which are often disruptive and infrequent, in-situ image data offers a non-invasive means to monitor key physical changes in real time. However, despite this potential, there remains a lack of effective CV-based algorithms tailored for online forecasting in drying processes. Existing approaches typically rely on reducing high-dimensional image data into a few tabular features~\cite{keramat2021real}, which leads to substantial information loss and diminishes their utility for precise modeling. Moreover, the challenges of handling small, noisy datasets and segmenting subtle visual features have further limited the development of robust predictive frameworks~\cite{xu2023small}. As a result, in-situ image data remains underutilized, and no existing work has fully leveraged its potential for dynamic process control in online forecasting.

Another key research gap is the lack of effective multi-modal integration methods for enhancing prediction accuracy. Food drying generates heterogeneous data modalities, each providing complementary insights. While multi-modal fusion holds promise for improving model robustness~\cite{zhao2023ai}, current methods often convert diverse inputs into simplified tabular formats. This not only increases preprocessing complexity but also leads to substantial information loss and limits the system’s flexibility in handling varied data types~\cite{petrich2021multi, keramat2021real}. As a result, they struggle to meet the demands of real-time, in-situ forecasting. For instance, although video data offers rich spatial-temporal information and is widely used in fields like autonomous driving~\cite{billard2019trends}, its integration with other modalities in food drying remains limited due to format mismatches and weak cross-modal feature extraction~\cite{yazici2018intelligent}. These demand a more generalizable fusion framework that uses modality-specific networks to directly process heterogeneous inputs and support robust, real-time predictions~\cite{li2025multi}.

\begin{figure}[H]
    \centering
    \includegraphics[width=1\columnwidth]{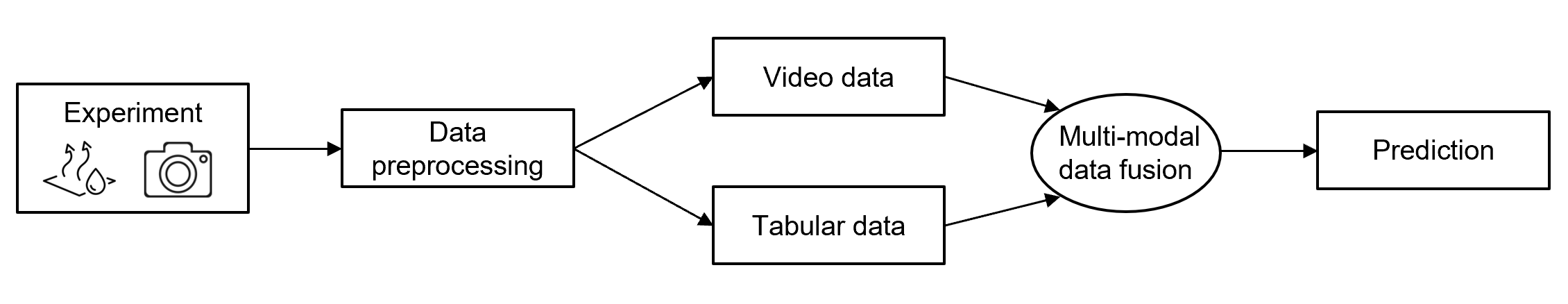}
    \caption{Schematic of the multi-modal real-time forecasting of cookie drying readiness framework.}
    \label{fig1}
\end{figure}

In this study, we propose an end-to-end multi-modal method for online readiness forecasting of food drying, which is illustrated by Figure~\ref{fig1}. Our approach leverages an end-to-end encoder-decoder network architecture that consists of modality-specific encoding networks to generate embeddings for each data modality without pre-transformation, and a transformer-based decoder to extract from multi-modal embeddings. We present a case study focused on predicting drying readiness using in-situ video data and process parameters. The proposed model is validated using leave-one-group-out cross-validation (LOGOCV). Our model achieves an average prediction error of only 15 seconds across the prediction window ranging from 120 to 10 seconds before readiness. Moreover, it shows a 65.69\% improvement over traditional fusion methods and an 11.30\% improvement over a video-only model. Our method is generalizable across various modality configurations and online monitoring applications.

The main contributions of our work are:

\begin{itemize}
    \item We propose a novel modality-flexible multi-modal fusion network for online forecasting of drying readiness. To the best of our knowledge, this study is among the first to directly take in-situ video data as model input in the drying area. 
    \item We demonstrate that multi-modal fusion provides a significant improvement in prediction accuracy, thereby laying the foundation for future online decision-making research in industrial drying.
    \item The effectiveness of various video encoders in model multi-modal fusion model is systematically investigated, which reveals the correlations between model complexity and prediction accuracy for limited, unbalanced data. These insights will be valuable for future multi-modal fusion research.

\end{itemize}

The remainder of this paper is organized as follows. Section 2 describes the dataset for the case study. Section 3 details the methodology, including the problem statement and detailed model architecture. Section 4 presents the data splitting strategy, training configuration, results, and comparative analysis with the baseline model, along with an ablation study. Section 5 discusses model performance under various configurations. Finally, Section 6 concludes the paper and outlines directions for future research.

\section{Dataset}


\subsection{Design of experiments and data collection}


We select sugar cookie drying as the case study, which transforms raw cookie doughs into edible final products. Typically, cookie drying takes place in high-temperature, sealed environments over a short period \cite{ergun2010moisture}. Throughout this process, the cookie’s texture, color, and shape undergo significant changes \cite{farzad2021drying}. However, variations in process parameters and inherent sample differences can lead to over-drying and under-drying. Over-drying causes excessive hardness, cracking, or undesirable texture, whereas under-drying leaves residual moisture that compromises structural integrity and shelf stability \cite{hsu1992determining}. Given these challenges, in-situ monitoring of sugar cookie drying is essential for determining the optimal stopping point, thereby ensuring ideal drying conditions, preventing defects, and maintaining consistent product quality.

In this research, raw samples are Pillsbury sugar cookie dough. Each raw piece weighs approximately 18.5 g and is shaped into a round form with a diameter of 60–70 mm and a thickness of 6–7 mm. After drying, the final weight is reduced to 15–16 g, with an expanded diameter of 80–90 mm and a thickness of 8–10 mm.

\begin{figure}[H]
    \centering
    \includegraphics[width=0.5\columnwidth]{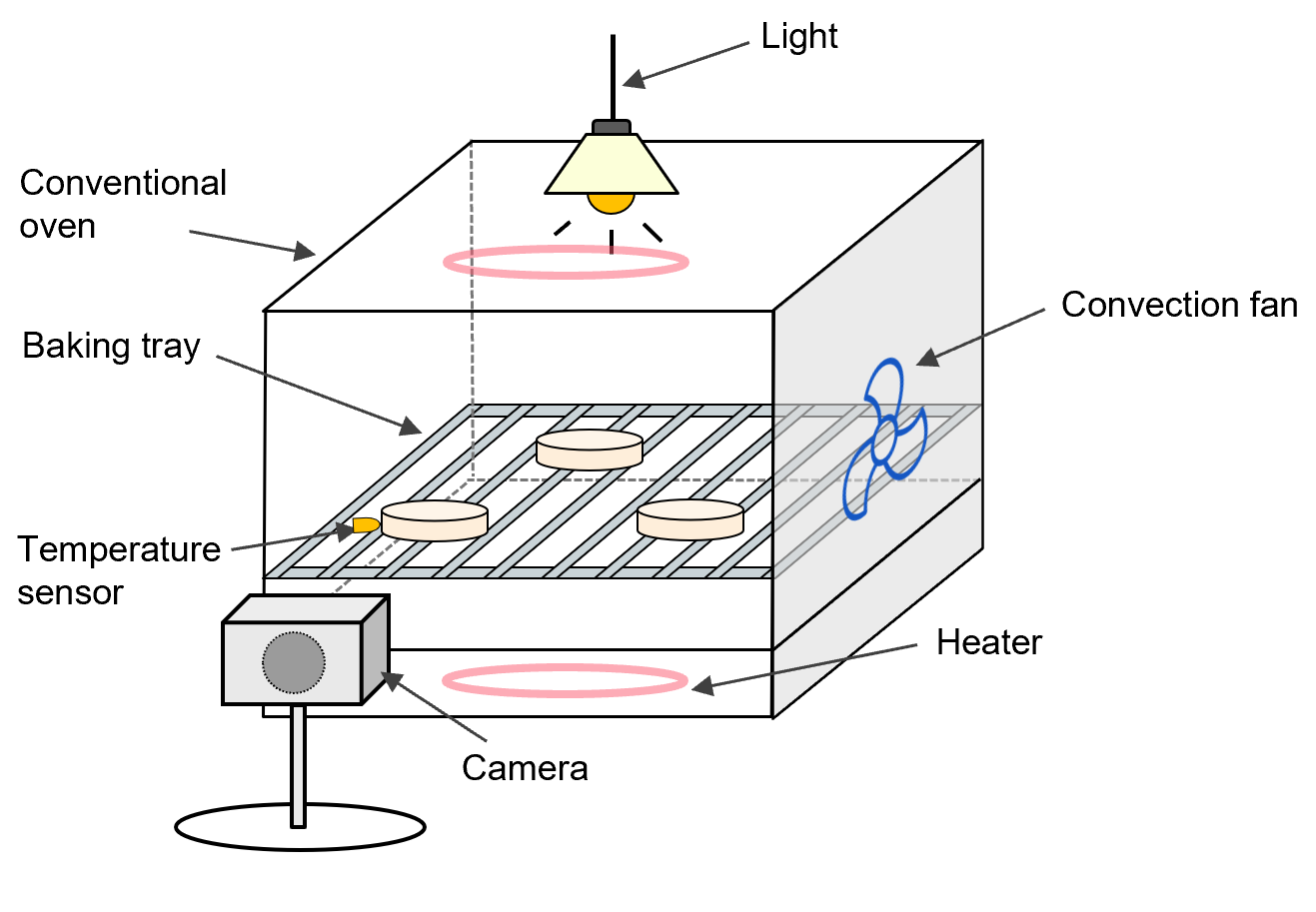}
    \caption{Schematic setup of the cookie drying experiment.}
    \label{fig3}
\end{figure}

The dryer is a Nuwave Smart Oven under controlled temperature and airflow conditions. Figure~\ref{fig3} illustrates the schematic setup of experiments. The oven operates in bake mode, utilizing top and bottom heating elements and a convection fan to circulate hot air. To enable real-time monitoring without interrupting the drying process, we mount a camera to record in-situ cookie drying videos, capturing one frame every ten seconds. A 5000K LED light is placed above the oven to maintain stable lighting conditions. Since the camera's limitation in temperature-resistance, images are recorded externally through the oven’s see-through glass door at a tilted angle, introducing visual noise from glass reflections and internal lighting. Each experiment involves drying three cookies simultaneously, meaning that each frame captures three samples at a time. This experimental setup is easily repeatable.

The cookies are dried under four temperature levels ($350^\circ\mathrm{F}$, $375^\circ\mathrm{F}$, $385^\circ\mathrm{F}$, and $400^\circ\mathrm{F}$) and two air velocity conditions (1000 revolutions per minute (RPM) and 3000 RPM fan speed), resulting in eight drying environmental conditions. To maintain experimental consistency, the oven is preheated to ensure the chamber temperature and convection fan have stabilized at the target conditions. For each condition, three independent experiments are conducted, with three cookies dried simultaneously per experiment, leading to a total of 72 drying batches. Drying durations are 9 minutes per batch, ensuring that all cookies are either fully baked or overbaked to include the 'ready' moment. The oven’s built-in thermometer continuously monitors the internal chamber temperature during the drying process. 


We define the `ready' moment as the point when the cookie is fully baked and requires no further drying. At this stage, the reduction in moisture vapor release causes a sharp change in chamber temperature. The 'ready' moment is identified through continuous monitoring of the oven’s built-in thermometer and validated by two experienced cookie bakers to ensure accuracy~\cite{vance1996rate}. Since cookies in the same drying batch are exposed to identical environmental conditions with minimal variation in dough characteristics, we assume they reach the ready moment simultaneously. The ready moment is recorded for each drying process and annotated in the corresponding video data. We calculate the remaining drying time per frame leading up 10 seconds to the ready moment. Specifically, each image frame occurring before the ready moment is assigned a ‘time-to-ready’ label, indicating the estimated time remaining until fully baked. Our case study focuses on forecasting the time-to-ready value at each timestamp.


\subsection{Video data preparation}

\begin{figure}[H]
    \centering
    \includegraphics[width=0.7\columnwidth]{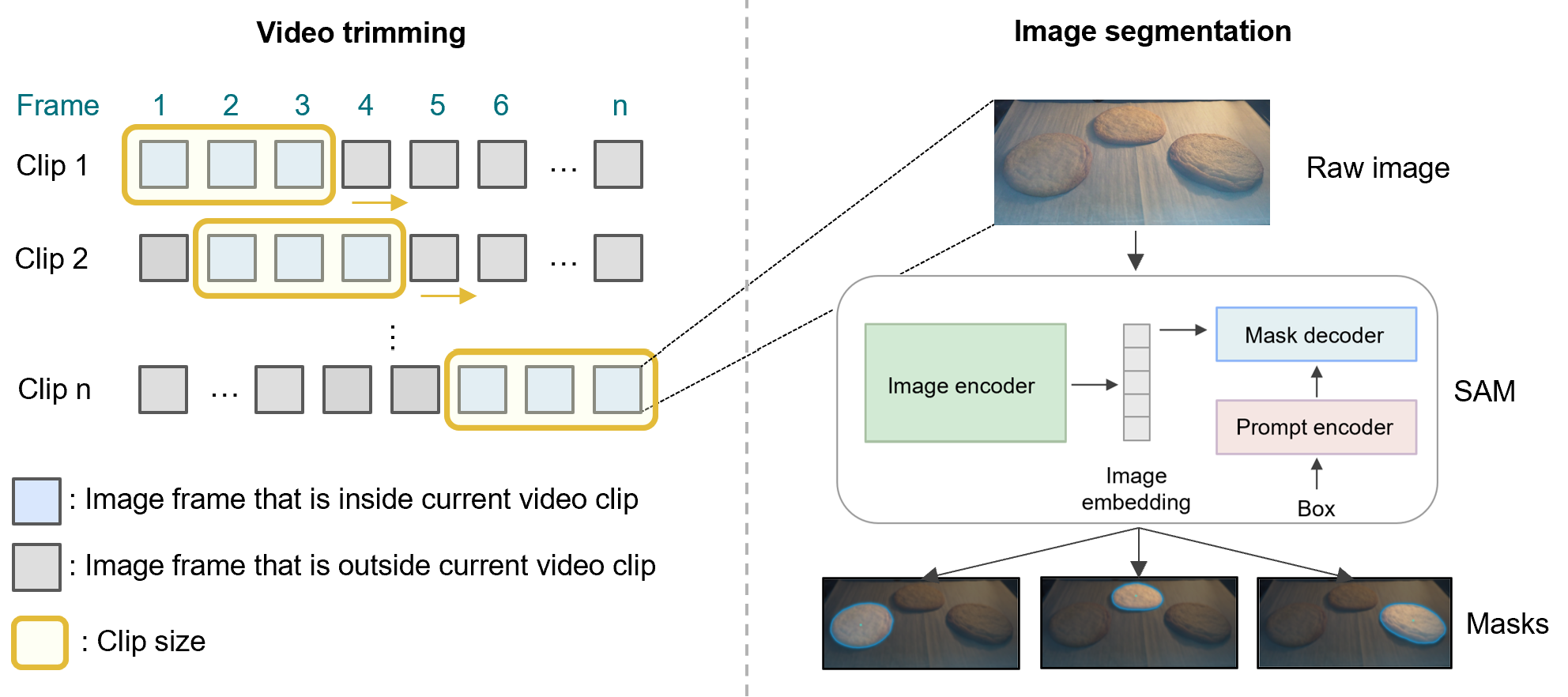}
    \caption{Video data preparation process.}
    \label{fig4}
\end{figure}

The preparation of the collected in-situ video data consists of two steps: video trimming and image segmentation. Figure~\ref{fig4} illustrates the video data preparation process. We apply a trimming process to extract short video clips as standardized inputs for model training. Using a sliding-window approach, we divide the full-length video into overlapping segments, each containing a fixed number of frames. This method ensures that each video clip captures cookies’ spatial changes over time while preserving both frame-specific and temporal information. Additionally, this trimming process increases the number of training and evaluation samples, enhancing model robustness and reducing computational costs compared to processing full-length videos.

Since each image frame contains three cookies, we employ the Segment Anything Model (SAM) for automated segmentation. SAM is a state-of-the-art foundation model designed for efficient object masking~\cite{kirillov2023segment}. Utilizing a pretrained image encoder and mask decoder, SAM generates precise masks for each cookie in milliseconds level, guided by box mask thresholds as prompts, including mask size, aspect ratio, and mask quality constraints. 

As all images are analyzed under consistent acquisition conditions, we do not apply standardization to pixel-wise RGB color values or the shapes of the cookies to simplify pre-processing procedures. Instead, we leverage the differences within and between frames, which underscores the necessity of employing video data over static images for real-time forecasting of food drying processes. 

\begin{figure}[H]
    \centering
    \includegraphics[width=0.7\columnwidth]{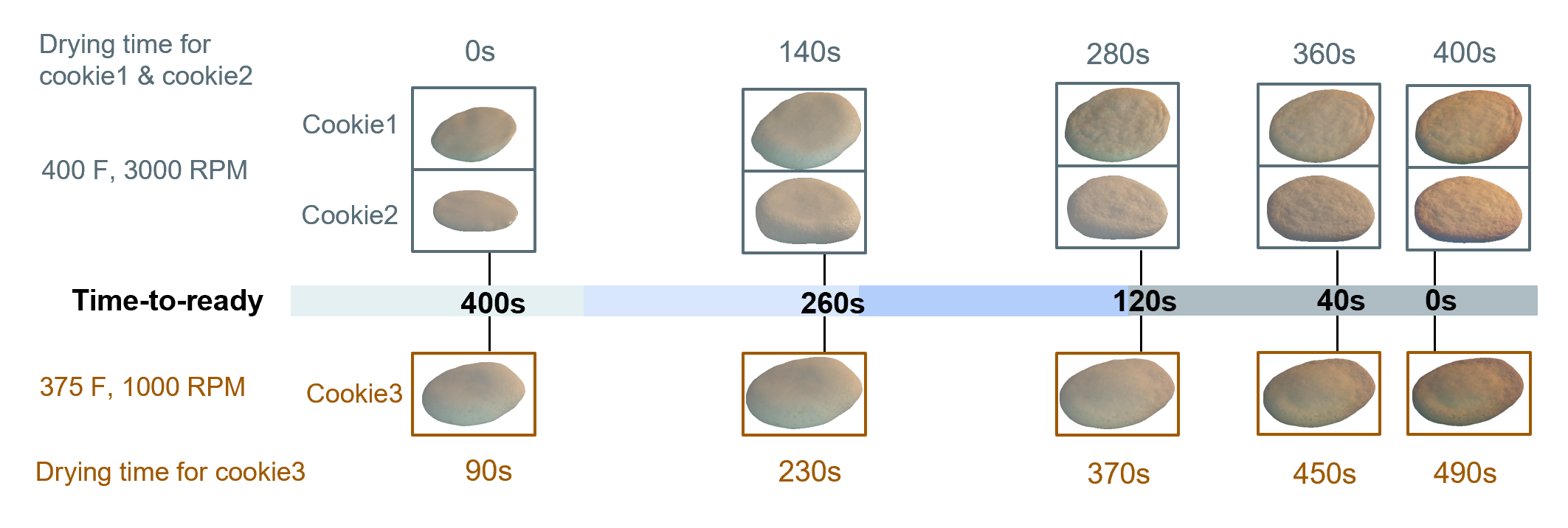}
    \caption{Examples of cookies at different time-to-ready moments.}
    \label{fig5}
\end{figure}

Figure~\ref{fig5} presents examples of cookies at different time-to-ready moments under various environmental conditions, as well as cookies exposed to the same environmental conditions but displaying different sample characteristics. Throughout each cookie’s drying process, we observe gradual changes in shape, texture, and color. In comparing Cookie 1 and Cookie 3 under different environmental conditions, we find that their appearances differ markedly at the same time-to-ready moment, demonstrating the impact of environmental drying parameters. Conversely, Cookie 1 and Cookie 2, which dry under identical environmental conditions, show similar drying stages but still display distinct variations in shape, texture, and color due to inherent differences in sample characteristics. These findings validate that cookie drying quality is impacted by both environmental settings as well as sample variances.  


\section{Methodology}

\subsection{Problem statement}

Our goal is to learn a mapping function:

\begin{equation}
    f: \left( \mathbf{X}_v, \mathbf{X}_t \right) \rightarrow y,
\end{equation}
where $\mathbf{X}_{v}$ and $\mathbf{X}_{t}$ are different data modalities, specifically, \( \mathbf{X}_{t} \in \mathbb{R}^{a} \) represents the tabular data containing \( a \) numerical process parameters (e.g., drying temperature, air velocity); and \( \mathbf{X}_{v} \in \mathbb{R}^{b \times C \times H \times W} \) denotes the in-situ video data, with \( b \) being the number of frames, \( C \) being the number of channels per frame, and \( H \) and \( W \) being the height and width of each frame, respectively. \( y \) represents the ground truth value of target variable time-to-ready. 

Therefore, we formulate the problem as and optimization of: 
\begin{equation}
\Theta = \{\Theta_t, \Theta_r, \Theta_g, \Theta_d, \Theta_o\},
\end{equation}

by minimizing:
\begin{equation}
\mathcal{L}_{\text{total}} = \frac{1}{N} \sum_{i=1}^N \mathcal{L}(y_i, \hat{y}_i) + \lambda \|\Theta\|^2,
\end{equation}
where $\theta$ is represents five layer-specific learnable parameters; $N$ is the number of samples; $\lambda$ is regularization constant; and $\mathcal{L}(y, \hat{y})$ is the Smooth L1 loss given by:
\begin{equation}
\mathcal{L}(y, \hat{y}) =
\begin{cases}
0.5 \, (y - \hat{y})^2, & \text{if } |y - \hat{y}| < 1, \\
 |y - \hat{y}| - 0.5 , & \text{otherwise,}
\end{cases}
\end{equation}
where $\hat{y} \in \mathbb{R}$ is the one-dimensional predicted output.

\subsection{Model architecture}

\begin{figure}[H]
    \centering
    \includegraphics[width=1\textwidth]{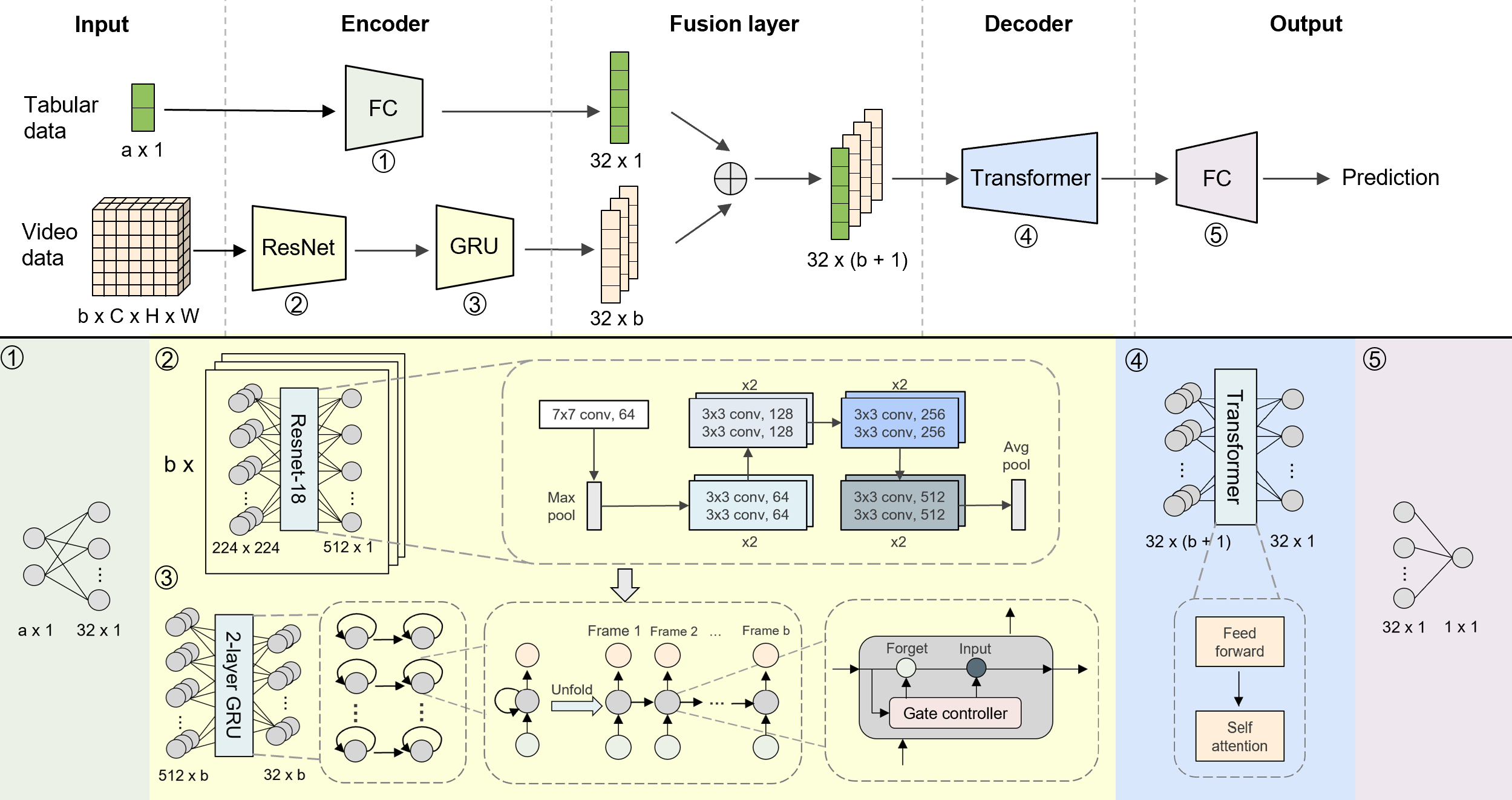}
    \caption{Overall architecture of multi-modal fusion model for online forecasting.}
    \label{fig:multi-modal-architecture}
\end{figure}

Figure~\ref{fig:multi-modal-architecture} illustrates the overall architecture of our multi-modal fusion model for online forecasting. The model follows a concatenated encoder-decoder structure, processing both tabular and video data as inputs. It consists of four key layers: encoder, fusion, decoder, and output layer, which work together to generate the final prediction.


\textbf{Encoder layer.} The encoder layer processes each modality independently, using specialized ML models to extract embedded representations while preserving structural integrity. For tabular data, a two-layer fully connected network (FC) transforms it to a 32 x 1 vector, aligning it with the video features for downstream fusion. This transformation is defined as:
\begin{equation}
\begin{aligned}
\mathbf{h}_t &= f_{\text{FC}}(\mathbf{X}_{t}; \Theta_{t}) \\
&= \mathbf{W}_{t}^{(2)} \sigma( \mathbf{W}_{t}^{(1)} \mathbf{X}_{t} + \mathbf{b}_{t}^{(1)}) + \mathbf{b}_{t}^{(2)} \\
&\in \mathbb{R}^{32 \times 1}
\end{aligned}
\end{equation}
where $\Theta_t = \left\{ \mathbf{W}_t^{(1)}, \mathbf{b}_t^{(1)}, \mathbf{W}_t^{(2)}, \mathbf{b}_t^{(2)} \right\}$ represents the set of all trainable parameters of the fully connected network at time step \( t \), which includes: \( \mathbf{W}_t^{(1)} \in \mathbb{R}^{32 \times a} \), weight matrix for the first layer; \( \mathbf{b}_t^{(1)} \in \mathbb{R}^{32 \times 1} \), bias vector for the first layer; \( \mathbf{W}_t^{(2)} \in \mathbb{R}^{32 \times 32} \), weight matrix for the second layer; \( \mathbf{b}_t^{(2)} \in \mathbb{R}^{32 \times 1} \), bias vector for the second layer. The activation function \( \sigma \) used in this network is the Rectified Linear Unit (ReLU), which is defined as:
\begin{equation}
    \sigma(x) = \max(0, x)
\end{equation}
ReLU introduces non-linearity to the model, allowing it to learn complex patterns by setting negative values to zero while keeping positive values unchanged.

The video data is treated as a time-series of image frames, with spatial features extracted using a pretrained ResNet-18 and temporal dependencies (exist exclusively in video data) captured by a two-layer Gated Recurrent Unit (GRU). ResNet-18 is a convolutional neural network based on residual learning. It extracts high-dimensional spatial features from each frame, while pretraining enhances feature extraction on limited or imbalanced datasets~\cite{he2016identity}. Two-layer GRU is a type of Recurrent Neural Network architecture that utilizes gating mechanisms to capture short- and long-term dependencies across frames in sequential data~\cite{chung2014empirical}. The encoded video representation is structured as 32 x b, where b is the number of frames per video clip. A comparative analysis of b value selection and encoder model choices are presented in Section 5.1 and 5.2. 

With 
\begin{equation}
\mathbf{F}_v = f_{\text{ResNet}}(\mathbf{X}_v; \Theta_r) \in \mathbb{R}^{b \times F},
\end{equation}
where $\mathbf{F}_v$ is the dimensionality of the ResNet output per frame, the GRU processes these extracted features as:
\begin{equation}
\mathbf{h}_v = f_{\text{GRU}}(\mathbf{F}_v; \Theta_g) \in \mathbb{R}^{32 \times b}
\end{equation}

\textbf{Fusion layer.} We concatenate the tabular embedding $\mathbf{h}_{t} \in \mathbb{R}^{32 \times 1}$ with the video embedding $\mathbf{h}_{v}\in \mathbb{R}^{32 \times b}$ along the temporal dimension by treating the tabular embedding as an additional “frame”. Specifically: 
\begin{equation}
\mathbf{h}_{\text{fusion}} = \text{concat}(\mathbf{h}_t, \mathbf{h}_v) \in \mathbb{R}^{32 \times (b + 1)}.
\end{equation}

\textbf{Decoder layer.} The decoder layer processes the fused embedding, captures underlying patterns and relationships to distill the information into a dense, one-dimensional vector for final predictions. The encoder module of a Transformer serves as the decoder. Transformer processes embedded data as parallel tokens and extracts relationships via self-attention mechanisms~\cite{vaswani2017attention}. This enables efficient handling of long-range dependencies, allowing the model to attend to relevant context across the fused embeddings. The self-attention mechanism further adapts weights dynamically, enhancing robustness in sequential tasks. The self-attention is computed as:

\begin{equation}
\text{Attention}(\textbf{Q}, \textbf{K}, \textbf{V}) = \text{softmax}\!\Bigl(\frac{\textbf{Q}\textbf{K}^\top}{\sqrt{d_k}}\Bigr)\textbf{V},
\end{equation}
where $\textbf{Q}$, $\textbf{K}$, and $\textbf{V}$ are linear projections of \(h_{\text{fusion}}\); and \(d_k\) is the dimension of the key vectors. 

The decoder takes a 32 x (b + 1) embedded representation as input and extracts a 32 x 1 output, defined as:
\begin{equation}
\mathbf{h}_{\text{d}} = f_{\text{Transformer}}(\mathbf{h}_{\text{fusion}}; \Theta_d) \in \mathbb{R}^{32 \times 1}.
\end{equation}

\textbf{Output layer.} A two-layer FC network maps the transformed features to the final prediction, defined as:
\begin{equation}
\hat{y} = f_{\text{out}}(\mathbf{h}_{\text{d}}; \Theta_o).
\end{equation}

\subsection{Baseline model}

\begin{figure}[H]
    \centering
    \includegraphics[width=0.5\columnwidth]{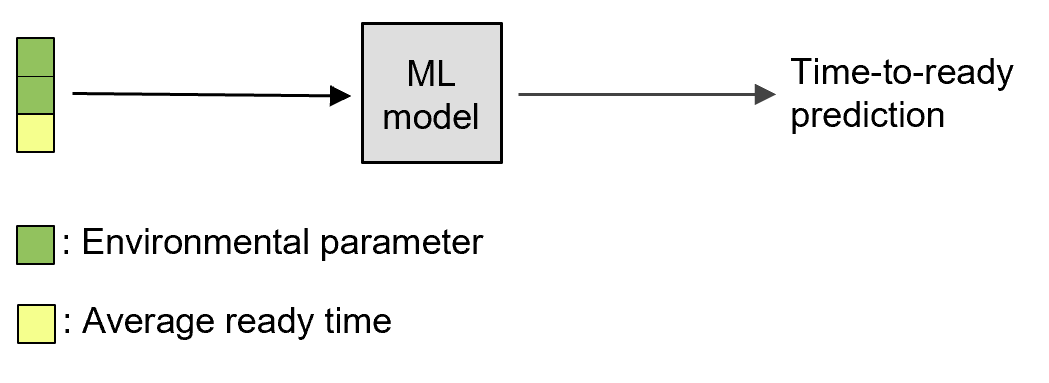}
    \caption{Architecture of baseline model.}
    \label{fig7}
\end{figure}

We compare our proposed methodology against a baseline model that employs a traditional single-modality fusion approach, where all data sources are converted into a single tabular format before being processed by an ML model. This widely used fusion strategy represents a classic approach in multi-source data integration. Figure~\ref{fig7} illustrates the baseline setup.

Specifically in our case study, the baseline model fuses process parameters (temperature and air velocity) with the average ready time observed under each process parameter combination. This aggregated tabular representation is then used to predict the ready time under a given drying condition. Given the simple input-output relationship, we evaluate multiple ML models for the baseline approach and find that a linear regression model provides the best fit. The time-to-ready estimation is then derived as the predicted ready time minus current drying duration. This method requires continuous tracking of drying duration, thus it's not online monitoring. 


\subsection{Ablation study}

\begin{figure}[H]
    \centering
    \includegraphics[width=0.8\columnwidth]{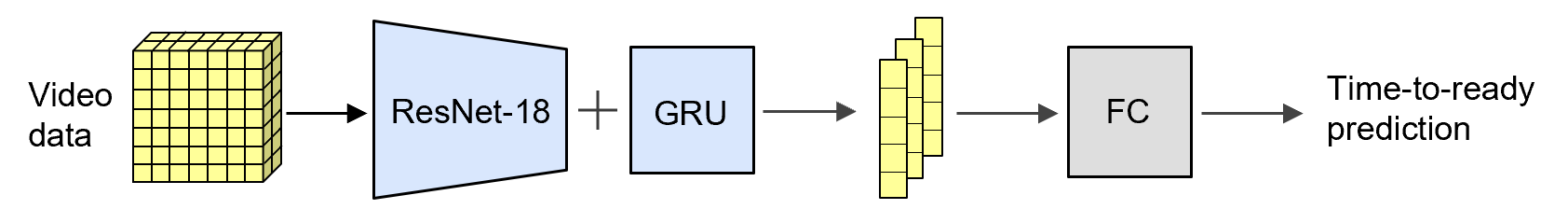}
    \caption{Architecture of ablation study model.}
    \label{fig8}
\end{figure}

Figure~\ref{fig8} illustrates the structure of the ablation study of a video-only model. Since the prediction task primarily relies on video data, we compare our multi-modal fusion method against a single-modality model that uses only video data to demonstrate the effectiveness of multiple modalities in time-to-ready prediction. Notably, using video data for real-time prediction is novel and has not been explored in previous studies. We conduct this comparison to justify the need for multi-modal data fusion in time-to-ready prediction. For a fair comparison, the encoder structure remains identical to that of the multi-modal model. Since only a single modality is used, parallel similarity calculations via the Transformer are unnecessary, the decoder is replaced with a FC for final prediction.

\section{Results}

\subsection{Dataset splitting}

In the food drying industry, datasets are often limited in size and variability. In many industrial applications, predictive models must generalize to unseen drying conditions that differ from those in the training set. Therefore, dataset splitting should account for non-i.i.d. (independent and identically distributed) conditions to ensure a realistic evaluation setting~\cite{eslaminia2024federated}.

\begin{figure}[H]
    \centering
    \includegraphics[width=0.6\columnwidth]{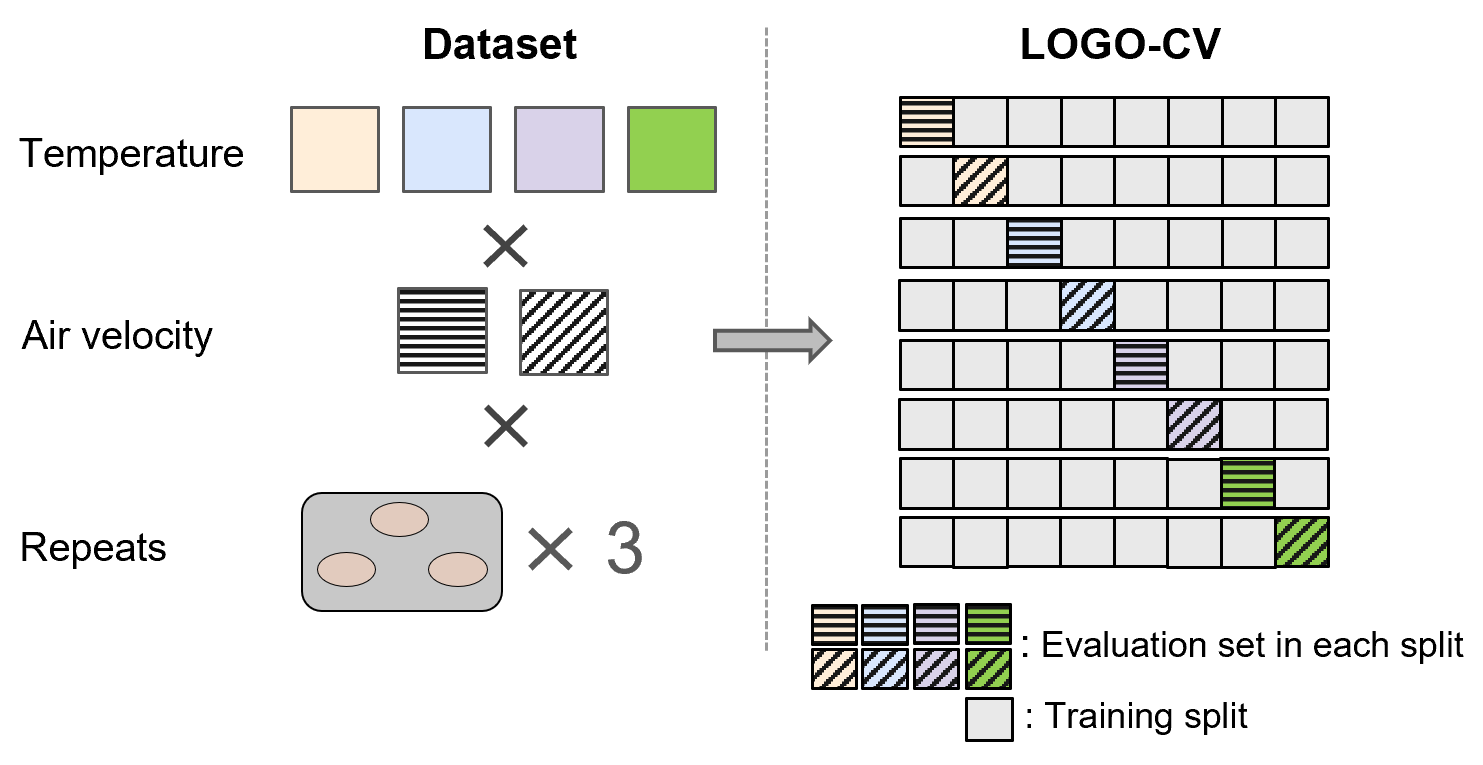}
    \caption{Illustration of the LOGOCV data splitting strategy.}
    \label{fig6}
\end{figure}

However, standard cross-validation assumes an i.i.d. dataset, where training and evaluation sets are randomly split~\cite{cody2020field}. This approach does not reflect real-world drying scenarios, where process parameters remain fixed within each drying batch, with sample characteristics introduce uncertain variances. As a result, identical drying process parameters in both training and evaluation sets can dominate and lead to data leakage. To ensure robust model evaluation, we employ a LOGOCV strategy, which partitions the dataset based on process parameter combinations rather than random sampling. As shown in Figure~\ref{fig6}, temperature and air velocity are controllable and remain fixed during drying, while sample characteristics are inherently variable. Specifically, we define eight unique process parameter combinations, corresponding to eight-fold cross-validation. Each fold trains on seven process parameter combinations and evaluates on the remaining unseen combination, ensuring the model is tested on conditions it has never encountered during training. 

The LOGOCV strategy typically results in larger prediction errors compared to standard cross-validation. However, a model that performs well under LOGOCV demonstrates strong generalizability, making it more reliable for real-world applications.

\subsection{Training configuration and performance metrics}


We forecast the time-to-ready range from 120 seconds to 10 seconds before the ready moment. This prediction window provides sufficient time for in-situ human response time while remaining computationally efficient and relevant to industrial applications. Additionally, visual changes in cookies become less pronounced in this late-stage drying phase, making manual estimation by experienced workers more difficult, thus a precise prediction model is critical.

Each input video clip consists of seven consecutive frames, capturing recent time-series variations in drying dynamics. The rationale behind this frame selection is discussed in Section 5.1. Therefore, our task is to predict the time-to-ready value at the current moment using the most recent seven frames and corresponding tabular data, given a non-i.i.d. dataset split based on process parameters.

Table~\ref{t1} presents the hyperparameters used during model training. The hyperparameters and model parameters are determined through a grid search approach to ensure optimal performance.

\begin{table}[H]
\centering
\caption{Hyperparameters for model training.}
\label{t1}
\arrayrulecolor{black}
\begin{tabular}{ l l } 
 \hline
 \textbf{Hyperparameter} & \textbf{Value} \\
 \hline
 Batch size  & 32 \\
 Learning rate & 0.0001 \\
 Number of epochs & 100 \\
 Optimizer & Adam \\
 Training metrics & Mean Absolute Error (MAE) \\
\hline
\end{tabular}
\end{table}

For both training and evaluation, we use mean absolute error (MAE) as the performance metric. MAE is calculated as:
\begin{equation}
\text{MAE} = \frac{1}{n} \sum_{i=1}^{n} | y_i - \hat{y}_i |,
\end{equation}
where n is the total number of samples, $y_i$ is the actual value of the i-th sample, $\hat{y}_i$ is the predicted value of the i-th sample, $| y_i - \hat{y}_i |$ is the absolute difference between the actual and predicted values at $i-th$ over $n$ timestamps. MAE treats all errors linearly, making it robust to outliers while offering better interpretability compared to alternative loss metrics~\cite{nie2010efficient}. Given the characteristics of our dataset, MAE provides an intuitive and reliable measure of prediction accuracy on each timestamp. A lower MAE means a better prediction accuracy and a better model performance. 

All models are trained on an NVIDIA GH200 GPU, which provides efficient computational resources for handling video-based deep learning models.

\subsection{Results and analysis}

Table~\ref{t2} presents the MAEs for the baseline model, video-only model, and our proposed multi-modal fusion model, along with the relative improvement of our approach compared to the other two methods. The baseline model performs significantly worse than both the video-only and multi-modal models. This may be caused due to baseline model relies solely on environmental parameters and the corresponding average ready time, failing to capture sample-specific variations and drying dynamics from in-situ data. Additionally, the baseline model depends heavily on experiences-based relationships. This limitation is particularly evident when the model encounters unseen drying conditions, where variations between the training and evaluation datasets lead to higher prediction errors.

The video-only model outperforms the baseline model, demonstrating the importance of extracting in-situ visual features. However, it still underperforms compared to our multi-modal fusion model, with an 11.03\% higher MAE. This performance gap suggests that while video data primarily reflects sample characteristics, environmental parameters remain essential for accurate prediction. The absence of environmental parameters in the video-only model likely reduces its ability to generalize across varying drying conditions, as it lacks additional contextual information to guide frame-to-frame learning during training. By incorporating environmental parameters, our multi-modal model effectively learns complex drying patterns, leading to more robust predictions.

Within the 120s to 10s range before readiness, our multi-modal model achieves an average MAE of approximately 15s. Physically, this means our model can predict the remaining drying time at any given moment with an average deviation of only 15 seconds. Moreover, compared to the baseline model, our approach eliminates the need for continuous tracking from the start of the drying process, allowing for more flexible and efficient monitoring from any current moment. 

\begin{table}[H]
\centering
\caption{Comparison of average MAE and MAE reduction between baseline and ablation study models with the proposed multi-modal model.}
\label{t2}
\arrayrulecolor{black}
\begin{tabular}{ l  l  l } 
 \hline
 \textbf{Method} & \textbf{MAE} & \textbf{MAE reduction} \\
 \hline
 Baseline & 44.92 & 65.69\% \\
 Video-only & 17.33 & 11.03\% \\
 Multi-modal & 15.41 & - \\
\hline
\end{tabular}
\end{table}

\begin{figure}[H]
    \centering
    \includegraphics[width=0.5\columnwidth]{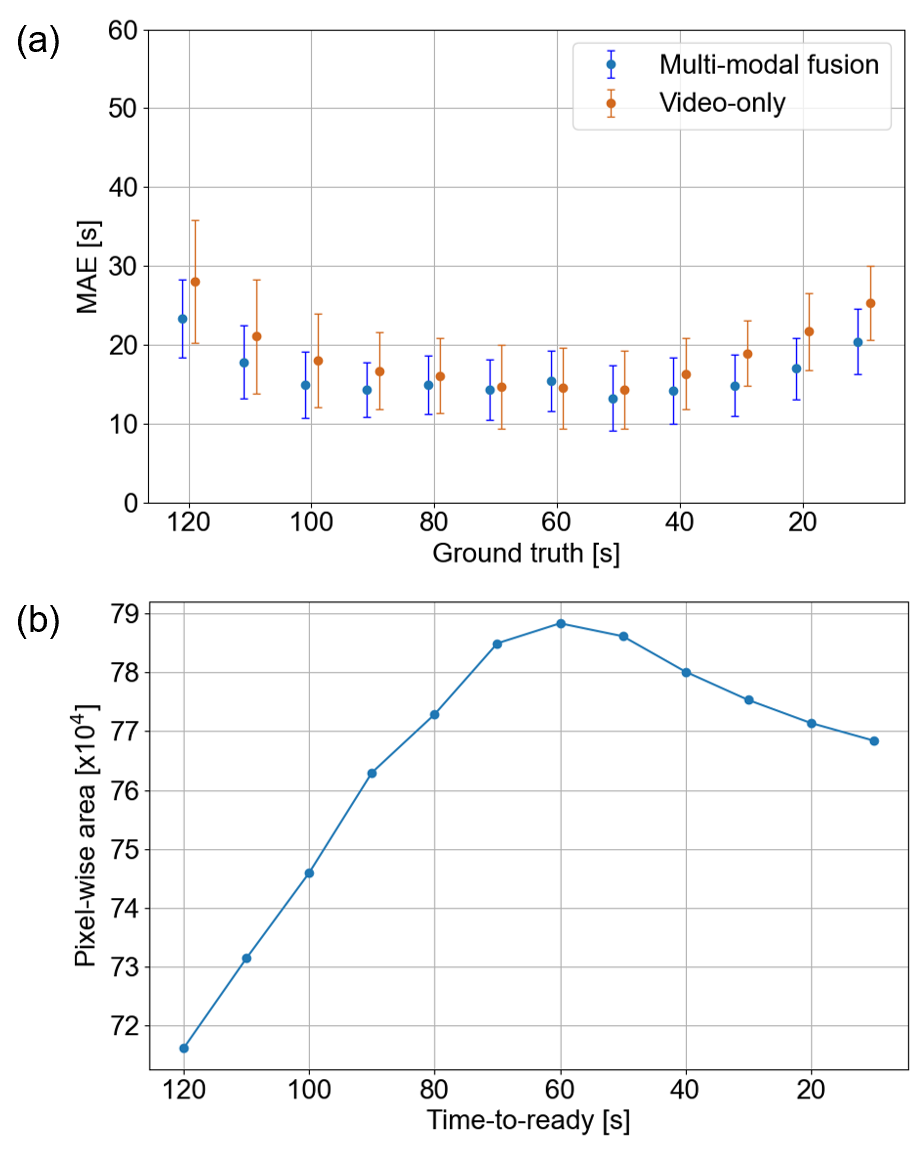}
    \caption{Values on each timestamp between 120s to 10s before readiness for (a) average MAE and its standard deviation on each timestamp for Video-only and Multi-modal methods and (b) average pixel-wise area.}
    \label{fig9}
\end{figure}

Figure~\ref{fig9} compares the MAE of the video-only and multi-modal fusion models across different time-to-ready moments. In Figure~\ref{fig9}(a), the dots represent the average MAE at each timestamp, while the error bars indicate standard deviation. The multi-modal fusion model achieves more stable predictions across all timestamps, with lower standard deviations, indicating higher reliability. Notably, the multi-modal model shows superior accuracy at both the earliest and latest drying stages compared to the video-only model. We observe that both models achieve their lowest MAE around 60 seconds before readiness. As illustrated in Figure~\ref{fig9}(b), this corresponds to the time window where cookies exhibit the largest shape, which may make their drying state more visually distinguishable. This suggests that shape variations play a critical role in the model’s ability to make accurate predictions, further reinforcing the advantage of integrating video data with environmental parameters.

\section{Discussion}

In this section, we conduct additional experiments to evaluate the impact of different model and variable selections on the performance of our multi-modal fusion framework. Specifically, Section 5.1 analyzes the effect of the number of input video frames on model performance and justifies our choice of seven frames; Section 5.2 evaluates the robustness and generalizability of our model by varying the number of training-evaluation set size, comparing multi-modal and single-modality models; Section 5.3 examines the impact of different video encoders, revealing the correlations of an encoder's complexity and accuracy for small, non-i.i.d. datasets.

\subsection{Selection of number of video frames}

We evaluate the optimal number of frames for input video clips by analyzing the trade-offs between accuracy and efficiency in our multi-modal fusion model. The accuracy is measured using average MAE across all timestamps, while efficiency is assessed based on the average inference time per batch.

\begin{figure}[h]
    \centering
    \includegraphics[width=0.5\columnwidth]{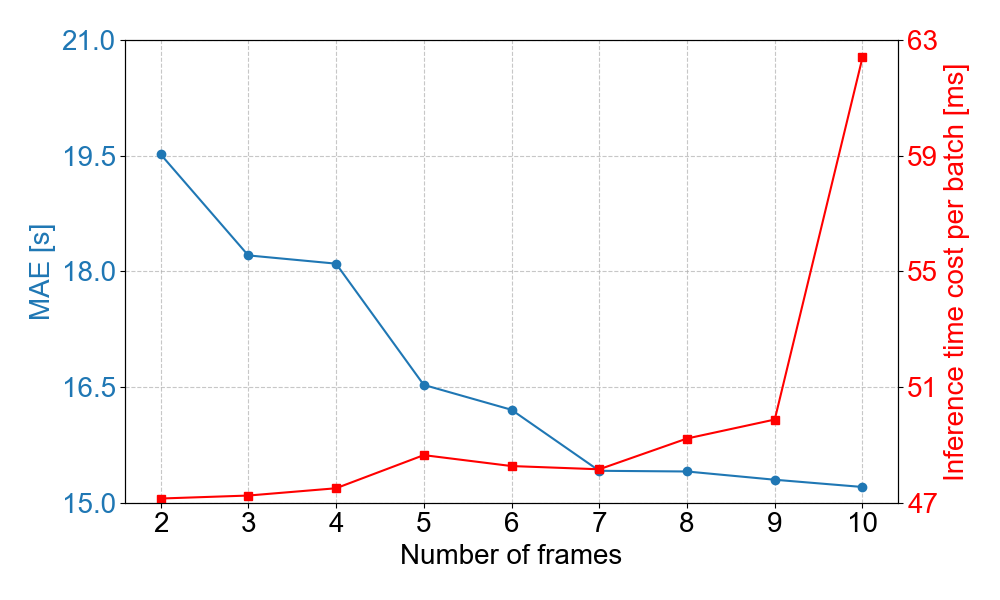}
    \caption{Accuracy-efficiency relationship for different frame counts.}
    \label{fig10}
\end{figure}

Figure~\ref{fig10} presents the accuracy-efficiency relationship for different frame counts, with the blue curve representing accuracy and the red curve representing inference time. As the number of frames increases, accuracy improves due to the additional temporal information captured. However, inference time also rises as processing more frames require greater computational resources. We observe that beyond seven frames, the improvement in accuracy diminishes, while the inference time increases sharply. Thus, we select seven frames per video clip as the optimal balance between accuracy and efficiency. This methodology for frame selection provides a generalizable approach for determining the appropriate input size in similar video-based monitoring tasks.

Furthermore, we compare the performance degradation (MAE increase) of our multi-modal model and a video-only model when the number of frames is reduced to 5, 3, and 1 (single image). Table~\ref{t3} summarizes the results.

\begin{table}[H]
\centering
\caption{MAE comparison of video-only and multi-modal method, and MAE reduction between the multi-modal over video-only method across different numbers of frames.}
\label{t3}
\arrayrulecolor{black}
\begin{tabular}{ l l l l } 
 \hline
 \textbf{Number of frames} & \textbf{Video-only} & \textbf{Multi-modal} & \textbf{MAE reduction} \\
 \hline
 7 & 17.325365 & 15.41358 & 11.03\% \\
 5 & 20.49349 & 16.52797 & 19.35\% \\
 3 & 23.49714 & 18.206897 & 22.51\% \\
 1 (image) & 31.76666 & 22.51177 & 29.13\% \\
\hline
\end{tabular}
\end{table}

As the number of frames decreases, both models experience a drop in accuracy. However, the video-only model exhibits a greater performance decline, indicating that environmental parameters in our multi-modal approach enhance model stability by compensating for the reduced temporal information. When using only one frame, both models perform significantly worse, reinforcing the necessity of temporal modeling in drying process prediction.

\subsection{Impact of training-evaluation data size on model robustness}

In most of of industrial drying cases, there are various of variables which cannot be fully considered in model training process. So models are desired to be robust with less training conditions and more complex evaluation sets. Our previous experiments follow an 8-fold LOGOCV approach, where each fold uses one environmental condition as the validation set while training on the remaining seven conditions. To further investigate model robustness, we increase the number of validation groups while reducing the available training data and observe the impact on performance. Specifically, we experiment with training set sizes of 7, 6, 5, and 4 groups, corresponding to evaluation set sizes of 1, 2, 3, and 4 groups, respectively. This setup maintains dataset consistency across both the multi-modal and video-only models for fair comparison. 

Table~\ref{t4} shows that as training data decreases and evaluation set increases, prediction accuracy degrades. However, the multi-modal model exhibits greater resilience to these changes, as evidenced by its increasing percentage improvement over the video-only model. This improvement may be attributed to the environmental conditions provided by tabular data, which help the model differentiate between varying evaluation conditions more effectively. The additional contextual information enables the model to better capture and extract critical features, improving its overall robustness.

\begin{table}[H]
\centering
\caption{Comparison of video-only and multi-modal method, and MAE reduction between the multi-modal over video-only method across different numbers of training groups.}
\label{t4}
\arrayrulecolor{black}
\begin{tabular}{ l l l l } 
 \hline
 \textbf{No. of training groups} & \textbf{Video-only} & \textbf{Multi-modal} & \textbf{MAE reduction} \\
 \hline
 7 & 17.325365 & 15.41358 & 11.03\% \\
 6 & 22.97875 & 20.4172 & 11.15\% \\
 5 & 24.55741 & 20.91077 & 14.85\% \\
 4 & 27.20679 & 23.29230 & 14.39\% \\
\hline
\end{tabular}
\end{table}

These findings highlight the generalizability and robustness of our multi-modal fusion framework, particularly in data-scarce scenarios.

\subsection{Investigations of video encoders on model performance}

As our framework follows a flexible encoder-decoder architecture, we evaluate various video encoders to determine the most suitable approach for small, non-i.i.d. cookie drying datasets. Processing video data is inherently challenging, and the choice of video encoder is crucial for effectively extracting both spatial and temporal features, particularly in datasets with subtle variations between frames.

Moreover, most CV models are designed and trained on large, i.i.d. datasets, where distinguishing features are clear and well-separated, making them fundamentally different from our case study. These discrepancies necessitate empirical evaluation to identify an encoder that performs effectively in our small, non-i.i.d. data scenario.

We experiment with the following video encoding models, with varying working principles and model sizes:

\begin{itemize}
    \item VideoMAE is a masked autoencoder-based model that learns video representations through self-supervised learning. This is one of the state-of-the-art video processing methods~\cite{tong2022videomae}.
    \item TimeSFormer is a pretrained transformer-based model that captures both spatial and temporal dependencies using divided space-time attention~\cite{bertasius2021space}.
    \item Multi-Recurrent Neural Network (MRN) is a memory-augmented network designed to improve temporal coherence by refining past frame features~\cite{graves2007multi}.
    \item CNN is a standard convolutional network that processes individual frames independently without explicit temporal modeling~\cite{o2015introduction}.
    \item ResNet-18 is a lightweight residual network that enhances feature extraction through skip connections.
    \item ResNet-18 with GRU is a hybrid model that combines spatial feature extraction from ResNet-18 with temporal modeling from a GRU. Specifically, this is the model we use in our methodology.
\end{itemize}

\begin{figure}[H]
    \centering
    \includegraphics[width=0.6\columnwidth]{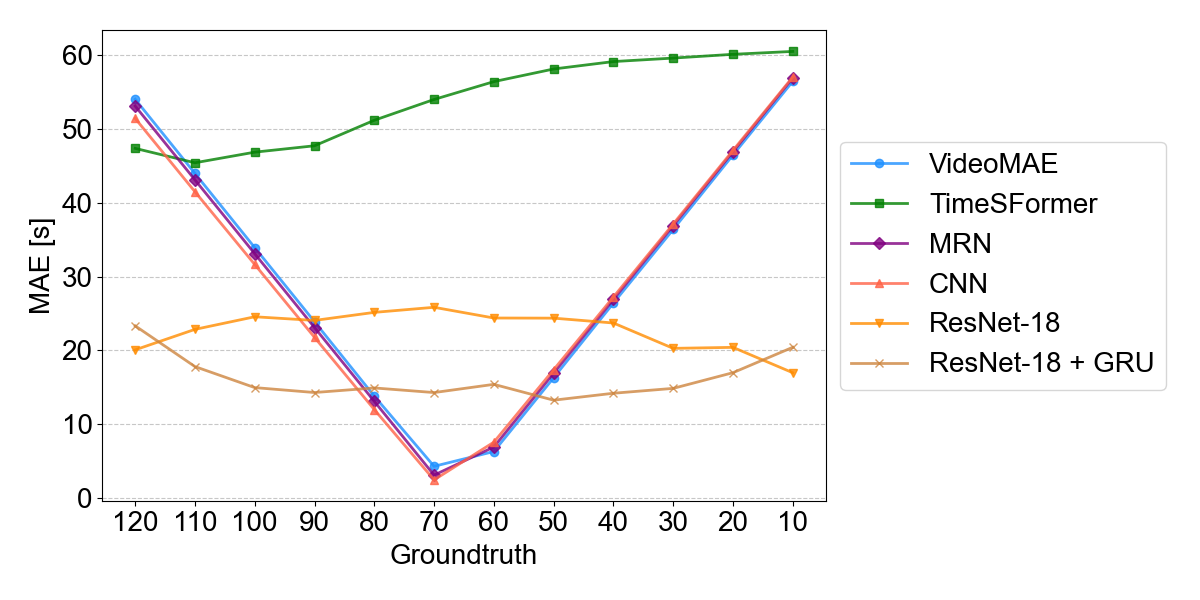}
    \caption{MAE curves across different timestamps for each video encoder.}
    \label{fig11}
\end{figure}

\begin{figure}[H]
    \centering
    \includegraphics[width=0.6\columnwidth]{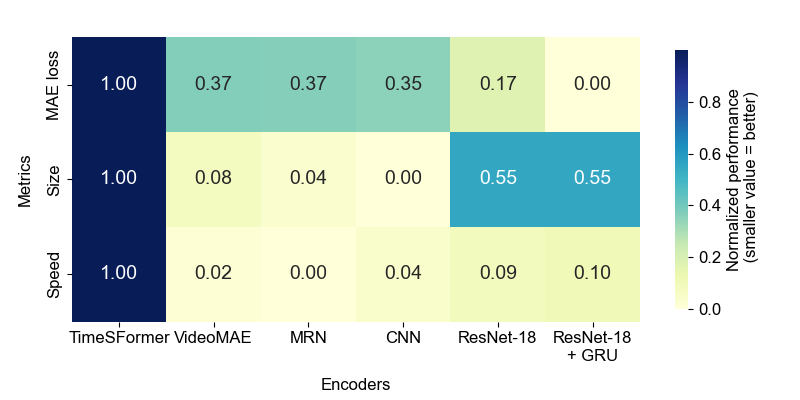}
    \caption{Normalized performance of video encoder models based on MAE, model size, and inference speed.}
    \label{fig12}
\end{figure}

Figure~\ref{fig11} presents the MAE curves across different timestamps for each video encoder. Additionally, Figure~\ref{fig12} visualizes the normalized performance of these encoder models based on MAE, model size, and inference speed, with lower values (lighter color) indicate better performance.

VideoMAE, CNN, and MRN exhibit similar and suboptimal predictions, with high variability in MAE across timestamps. As shown in Figure~\ref{fig11}, their predicted values remain nearly constant across different ground truth timestamps, indicating a failure to capture meaningful temporal dynamics. Meanwhile, Figure~\ref{fig12} shows that these three models have very small model sizes and fast computational speeds, suggesting that their limited complexity may hinder their ability to extract sufficient information from the video data. This implies that these models are too simple for the intricacies of our dataset, leading to reduced predictive accuracy.

TimeSFormer achieves diverse predictions across timestamps but performs poorly overall, indicated by the largest overall average MAE. It is likely due to high model complexity exceeding the capacity of our small, non-i.i.d. dataset. This results in overfitting and inefficiencies.

As a balanced compromise, ResNet-18 and ResNet-18 + GRU outperform all other encoders, achieving lower MAE and more stable predictions across all timestamps. Both models maintain a moderate model size and computational cost, making them more efficient than heavier architectures while still capturing essential spatial and temporal features. Among them, the hybrid ResNet-18 + GRU model delivers the best overall performance, reinforcing the importance of explicit temporal modeling in accurately predicting the drying process. Our findings suggest that ResNet-18 with GRU strikes an optimal balance—it is neither too simple to miss critical features nor too complex to overfit the small dataset, revealing the correlations between model complexity and efficiency, making it the most suitable choice for our case study.

These results highlight a key insight: state-of-the-art models designed for large, diverse datasets do not necessarily perform best on small, specialized industrial datasets. Selecting a model suited to dataset characteristics is critical for optimizing performance.

This investigation is among the first studies to systematically evaluate different ML architectures on a small, non-i.i.d. industrial dataset, demonstrating the importance of task-specific model selection. Our findings provide a framework for optimizing model architectures in similar industrial monitoring applications.

\section{Conclusion and future work}

In this paper, we present an end-to-end multi-modal real-time forecasting framework for in-situ quality prediction during the food drying process. This architecture consists of modality-specific encoders and a transformer-based decoder, effectively capturing the complex interplay between controllable environmental factors and intrinsic sample characteristics. Our approach predicts the in-situ time-to-ready value at each timestamp for cookie drying without interrupting the drying process by integrating in-situ video data and process parameters through an encoder–decoder architecture. Experimental results show that our method achieves an average discrepancy of only 15 seconds relative to the ground truth, demonstrating its accuracy and reliability, while achieving significant performance improvements of 65.69\% over traditional fusion methods and 11.3\% over a video-only model. The framework exhibits strong robustness, flexibility, and generalizability across various configurations and operational conditions, underscoring its potential as a valuable tool for real-time monitoring and quality control in industrial settings. Moreover, our findings indicate that the hybrid combination of ResNet-18 and a two-layer GRU model attains optimal prediction accuracy while maintaining a moderate model size and computational cost for small, non-i.i.d industrial datasets.

The findings of this paper highlight two important future directions to enhance the applicability of our framework in a wide range of industrial settings. First, the generalizability of our methodology can be explored across other industrial applications. This includes the incorporation of additional data modalities, such as real-time sensor measurements in ultrasonic welding. Second, we can explore transfer learning strategies to address data scarcity and process variability, including the systematic quantification of dataset similarities. 



\section*{Acknowledgements}

This study was financially supported by the U.S. Department of Energy, Office of Advanced Manufacturing under Award Number DE-EE0009125, the Massachusetts Clean Energy Center (MassCEC), and Center for Advanced Research in Drying (CARD). The views expressed herein do not necessarily represent the views of the U.S. Department of Energy or the United States Government.

This work used the DeltaAI system at the National Center for Supercomputing Applications through allocation MCH240099 from the Advanced Cyberinfrastructure Coordination Ecosystem: Services and Support (ACCESS) program, which is supported by National Science Foundation grants \#2138259, \#2138286, \#2138307, \#2137603, and \#2138296. And this research used the DeltaAI advanced computing and data resource, which is supported by the National Science Foundation (award OAC 2320345) and the State of Illinois. DeltaAI is a joint effort of the University of Illinois Urbana-Champaign and its National Center for Supercomputing Applications.






\bibliography{references}

\begin{thebibliography}{10}
\expandafter\ifx\csname url\endcsname\relax
  \def\url#1{\texttt{#1}}\fi
\expandafter\ifx\csname urlprefix\endcsname\relax\def\urlprefix{URL }\fi
\expandafter\ifx\csname href\endcsname\relax
  \def\href#1#2{#2} \def\path#1{#1}\fi

\bibitem{floros2010feeding}
J.~D. Floros, R.~Newsome, W.~Fisher, G.~V. Barbosa-C{\'a}novas, H.~Chen, C.~P. Dunne, J.~B. German, R.~L. Hall, D.~R. Heldman, M.~V. Karwe, et~al., Feeding the world today and tomorrow: the importance of food science and technology: an ift scientific review, Comprehensive Reviews in Food Science and Food Safety 9~(5) (2010) 572--599.

\bibitem{adnouni2023computational}
M.~Adnouni, L.~Jiang, X.~Zhang, L.~Zhang, P.~B. Pathare, A.~Roskilly, Computational modelling for decarbonised drying of agricultural products: Sustainable processes, energy efficiency, and quality improvement, Journal of Food Engineering 338 (2023) 111247.

\bibitem{farzad2021drying}
M.~Farzad, J.~Yagoobi, Drying of moist cookie doughs with innovative slot jet reattachment nozzle, Drying Technology 39~(2) (2021) 268--278.

\bibitem{joardder2015porosity}
M.~U. Joardder, A.~Karim, C.~Kumar, R.~J. Brown, Porosity: establishing the relationship between drying parameters and dried food quality, Springer, 2015.

\bibitem{arslan2024assessing}
A.~Arslan, {\.I}.~Aliba{\c{s}}, Assessing the effects of different drying methods and minimal processing on the sustainability of the organic food quality, Innovative Food Science \& Emerging Technologies (2024) 103681.

\bibitem{omolola2017quality}
A.~O. Omolola, A.~I. Jideani, P.~F. Kapila, Quality properties of fruits as affected by drying operation, Critical reviews in food science and nutrition 57~(1) (2017) 95--108.

\bibitem{defraeye2017impact}
T.~Defraeye, Impact of size and shape of fresh-cut fruit on the drying time and fruit quality, Journal of Food Engineering 210 (2017) 35--41.

\bibitem{li2025uncertainty}
S.~Li, A.~Malvandi, H.~Feng, C.~Shao, Uncertainty-aware constrained optimization for air convective drying of thin apple slices using machine-learning-based response surface methodology, Journal of Food EngineeringIn press (2025).
\newblock \href {https://doi.org/10.1016/j.jfoodeng.2025.112503} {\path{doi:10.1016/j.jfoodeng.2025.112503}}.

\bibitem{tian2023weldmon}
B.~Tian, A.~Eslaminia, K.-C. Lu, Y.~Wang, C.~Shao, K.~Nahrstedt, Weldmon: A cost-effective ultrasonic welding machine condition monitoring system, in: 2023 IEEE 14th Annual Ubiquitous Computing, Electronics \& Mobile Communication Conference (UEMCON), IEEE, 2023, pp. 0310--0319.

\bibitem{schmitt2020predictive}
J.~Schmitt, J.~B{\"o}nig, T.~Borggr{\"a}fe, G.~Beitinger, J.~Deuse, Predictive model-based quality inspection using machine learning and edge cloud computing, Advanced engineering informatics 45 (2020) 101101.

\bibitem{li2021human}
Q.~Li, K.~K. Ng, Z.~Fan, X.~Yuan, H.~Liu, L.~Bu, A human-centred approach based on functional near-infrared spectroscopy for adaptive decision-making in the air traffic control environment: A case study, Advanced Engineering Informatics 49 (2021) 101325.

\bibitem{qu2024development}
J.~Qu, L.~Cui, W.~Guo, L.~Bu, Z.~Wang, Development of a novel machine learning-based approach for brain function assessment and integrated software solution, Advanced Engineering Informatics 60 (2024) 102461.

\bibitem{zhao2023ai}
Y.~Zhao, Y.~Zhang, Z.~Li, L.~Bu, S.~Han, Ai-enabled and multimodal data driven smart health monitoring of wind power systems: A case study, Advanced Engineering Informatics 56 (2023) 102018.

\bibitem{jia2024hybrid}
S.~Jia, J.~Sun, A.~Howes, M.~R. Dawson, K.~C. Toussaint~Jr, C.~Shao, Hybrid physics-guided data-driven modeling for generalizable geometric accuracy prediction and improvement in two-photon lithography, Journal of Manufacturing Processes 110 (2024) 202--210.

\bibitem{meng2024meta}
Y.~Meng, Z.~Dong, K.-C. Lu, S.~Li, C.~Shao, Meta-learning-based domain generalization for cost-effective tool condition monitoring in ultrasonic metal welding, IEEE Transactions on Industrial Informatics (2024).

\bibitem{meng2023explainable}
Y.~Meng, K.-C. Lu, Z.~Dong, S.~Li, C.~Shao, Explainable few-shot learning for online anomaly detection in ultrasonic metal welding with varying configurations, Journal of Manufacturing Processes 107 (2023) 345--355.

\bibitem{zhao2021effects}
C.-C. Zhao, K.~Ameer, J.-B. Eun, Effects of various drying conditions and methods on drying kinetics and retention of bioactive compounds in sliced persimmon, Lwt 143 (2021) 111149.

\bibitem{el2023novel}
H.~S. El-Mesery, K.~Ashiagbor, Z.~Hu, W.~Alshaer, A novel infrared drying technique for processing of apple slices: Drying characteristics and quality attributes, Case Studies in Thermal Engineering 52 (2023) 103676.

\bibitem{mishra2023development}
N.~Mishra, S.~Jain, N.~Agrawal, N.~Jain, N.~Wadhawan, N.~Panwar, Development of drying system by using internet of things for food quality monitoring and controlling, Energy Nexus 11 (2023) 100219.

\bibitem{aghbashlo2014measurement}
M.~Aghbashlo, R.~Sotudeh-Gharebagh, R.~Zarghami, A.~S. Mujumdar, N.~Mostoufi, Measurement techniques to monitor and control fluidization quality in fluidized bed dryers: A review, Drying Technology 32~(9) (2014) 1005--1051.

\bibitem{chen2025reinforcement}
S.~Chen, H.~Yu, J.~Yagoobi, C.~Shao, Reinforcement learning constrained beam search for parameter optimization of paper drying under flexible constraints, arXiv preprint arXiv:2501.12542 (2025).

\bibitem{shang2023defect}
H.~Shang, C.~Sun, J.~Liu, X.~Chen, R.~Yan, Defect-aware transformer network for intelligent visual surface defect detection, Advanced Engineering Informatics 55 (2023) 101882.

\bibitem{keramat2021real}
M.~Keramat-Jahromi, S.~S. Mohtasebi, H.~Mousazadeh, M.~Ghasemi-Varnamkhasti, M.~Rahimi-Movassagh, Real-time moisture ratio study of drying date fruit chips based on on-line image attributes using knn and random forest regression methods, Measurement 172 (2021) 108899.

\bibitem{xu2023small}
P.~Xu, X.~Ji, M.~Li, W.~Lu, Small data machine learning in materials science, npj Computational Materials 9~(1) (2023) 42.

\bibitem{petrich2021multi}
J.~Petrich, Z.~Snow, D.~Corbin, E.~W. Reutzel, Multi-modal sensor fusion with machine learning for data-driven process monitoring for additive manufacturing, Additive Manufacturing 48 (2021) 102364.

\bibitem{billard2019trends}
A.~Billard, D.~Kragic, Trends and challenges in robot manipulation, Science 364~(6446) (2019) eaat8414.

\bibitem{yazici2018intelligent}
A.~Yazici, M.~Koyuncu, T.~Yilmaz, S.~Sattari, M.~Sert, E.~Gulen, An intelligent multimedia information system for multimodal content extraction and querying, Multimedia Tools and Applications 77 (2018) 2225--2260.

\bibitem{li2025multi}
S.~Li, C.~Shao, Multi-modal data fusion for moisture content prediction in apple drying, arXiv preprint arXiv:2504.07465 (2025).

\bibitem{ergun2010moisture}
R.~Ergun, R.~Lietha, R.~W. Hartel, Moisture and shelf life in sugar confections, Critical reviews in food science and nutrition 50~(2) (2010) 162--192.

\bibitem{hsu1992determining}
C.~C. Hsu, C.~A. Ward, R.~Pearlman, H.~Nguyen, D.~Yeung, J.~G. Curley, Determining the optimum residual moisture in lyophilized protein pharmaceuticals., Developments in biological standardization 74 (1992) 255--70.

\bibitem{vance1996rate}
W.~Vance, X.~Chen, S.~Scott, The rate of temperature rise of a subbituminous coal during spontaneous combustion in an adiabatic device: The effect of moisture content and drying methods, Combustion and Flame 106~(3) (1996) 261--270.

\bibitem{kirillov2023segment}
A.~Kirillov, E.~Mintun, N.~Ravi, H.~Mao, C.~Rolland, L.~Gustafson, T.~Xiao, S.~Whitehead, A.~C. Berg, W.-Y. Lo, et~al., Segment anything, in: Proceedings of the IEEE/CVF International Conference on Computer Vision, 2023, pp. 4015--4026.

\bibitem{he2016identity}
K.~He, X.~Zhang, S.~Ren, J.~Sun, Identity mappings in deep residual networks, in: Computer Vision--ECCV 2016: 14th European Conference, Amsterdam, The Netherlands, October 11--14, 2016, Proceedings, Part IV 14, Springer, 2016, pp. 630--645.

\bibitem{chung2014empirical}
J.~Chung, C.~Gulcehre, K.~Cho, Y.~Bengio, Empirical evaluation of gated recurrent neural networks on sequence modeling, arXiv preprint arXiv:1412.3555 (2014).

\bibitem{vaswani2017attention}
A.~Vaswani, Attention is all you need, Advances in Neural Information Processing Systems (2017).

\bibitem{eslaminia2024federated}
A.~Eslaminia, Y.~Meng, K.~Nahrstedt, C.~Shao, Federated domain generalization for condition monitoring in ultrasonic metal welding, Journal of Manufacturing Systems 77 (2024) 1--12.

\bibitem{cody2020field}
R.~A. Cody, S.~Narasimhan, A field implementation of linear prediction for leak-monitoring in water distribution networks, Advanced Engineering Informatics 45 (2020) 101103.

\bibitem{nie2010efficient}
F.~Nie, H.~Huang, X.~Cai, C.~Ding, Efficient and robust feature selection via joint l2, 1-norms minimization, Advances in neural information processing systems 23 (2010).

\bibitem{tong2022videomae}
Z.~Tong, Y.~Song, J.~Wang, L.~Wang, Videomae: Masked autoencoders are data-efficient learners for self-supervised video pre-training, Advances in neural information processing systems 35 (2022) 10078--10093.

\bibitem{bertasius2021space}
G.~Bertasius, H.~Wang, L.~Torresani, Is space-time attention all you need for video understanding?, in: ICML, Vol.~2, 2021, p.~4.

\bibitem{graves2007multi}
A.~Graves, S.~Fern{\'a}ndez, J.~Schmidhuber, Multi-dimensional recurrent neural networks, in: International conference on artificial neural networks, Springer, 2007, pp. 549--558.

\bibitem{o2015introduction}
K.~O'Shea, An introduction to convolutional neural networks, arXiv preprint arXiv:1511.08458 (2015).

\end{thebibliography}


\end{document}